\documentclass[letterpaper, 10 pt, conference]{ieeeconf}  % Comment this line out if you need a4paper

\IEEEoverridecommandlockouts                              % This command is only needed if 
\usepackage[utf8]{inputenc}
\let\labelindent\relax
\usepackage{amsfonts}       % An extended set of fonts for use in mathematics

\usepackage{wrapfig}
\usepackage{amsthm}
\usepackage{mathtools}      % Loads amsmath
\usepackage{amssymb}        % provides AMS fonts and symbols ex. \mathbb{...}
\usepackage{filecontents}
\usepackage{graphicx}       % enhanced support for graphics
\usepackage{marginnote}     % Add notes
\usepackage{marvosym}       % mvat
\usepackage{overpic}        % Add text on figures
\usepackage{tabularx}
\usepackage{subcaption}
\usepackage{cite}
\usepackage{color}
\usepackage[normalem]{ulem}
\usepackage{epstopdf}
\usepackage{enumitem}
\usepackage[normalem]{ulem}
\usepackage{lipsum}
\usepackage{url}
\usepackage[hidelinks]{hyperref}
\usepackage[dvipsnames]{xcolor}
\usepackage{diagbox}
\usepackage{multirow}
\usepackage{microtype}
\usepackage{xspace}

\makeatletter
\newif\if@restonecol
\makeatother

\usepackage[linesnumbered,ruled,vlined]{algorithm2e}%[ruled,vlined]{
\usepackage{algpseudocode}
\usepackage{amsmath}

\newif\ifdraft
\draftfalse
\drafttrue

\ifdraft

\usepackage{xcolor}
\usepackage{xargs} 
\usepackage[textsize=footnotesize]{todonotes}
\newcommandx{\sw}[2][1=]{\todo[linecolor=blue,
			backgroundcolor=blue!10,bordercolor=blue,#1]{Siwei: #2}}
\newcommandx{\tg}[2][1=]{\todo[linecolor=orange,
			backgroundcolor=orange!10,bordercolor=orange,#1]{Greaten: #2}}
\newcommandx{\jy}[2][1=]{\todo[linecolor=green,
			backgroundcolor=green!10,bordercolor=green,#1]{JJ: #2}}
\else
\newcommand{\sh}[1]{{}}
\newcommand{\tg}[1]{{}}
\newcommand{\jy}[1]{{}}
\fi

\newif\iftwocolumn
\twocolumntrue

\newtheorem{problem}{Problem}
\newtheorem{proposition}{Proposition}[section]
\newtheorem{lemma}{Lemma}[section]

\newtheorem{corollary}{Corollary}[section]
\newtheorem{theorem}{Theorem}[section]
\makeatletter
\def\subsubsection{\@startsection{subsubsection}% name
                                 {3}% level
                                 {\z@ \hspace*{1mm}}% indent (formerly \parindent)
                                 {0ex plus 0.1ex minus 0.1ex}% before skip
                                 {0ex}% after skip
                                 {\normalfont\normalsize\itshape}}% style
\makeatother

\newcommand{\mpp}{\textsc{MRPP}\xspace}
\newcommand{\ilp}{\textsc{ILP}\xspace}

\newcommand{\ecbs}{\textsc{ECBS}\xspace}
\newcommand{\rthddd}{\textsc{RTH3D}\xspace}
\newcommand{\rthdd}{\textsc{RTH2D}\xspace}

\newcommand{\rthdddlba}{\textsc{RTH3D-LBA}\xspace}

\newcommand{\rtpk}{\textsc{RTKD}\xspace}
\newcommand{\rtptwo}{\textsc{RT2D}\xspace}
\newcommand{\rtp}{\textsc{RT3D}\xspace}
\newcommand{\rtatwo}{\textsc{RTA2D}\xspace}
\newcommand{\rta}{\textsc{RTA3D}\xspace}
\title{Polynomial Time Near-Time-Optimal Multi-Robot Path Planning in Three Dimensions with Applications to Large-Scale UAV Coordination 
}

\newif\ifarxiv
\arxivtrue
\author{Teng Guo \qquad Si Wei Feng  \qquad Jingjin Yu% <-this % stops a space
\thanks{G. Teng, A. S. Feng and J. Yu are with the Department of 
Computer Science, Rutgers, the State University of New Jersey, Piscataway, NJ, USA. 
Emails: {\tt\small \{ teng.guo, siwei.feng, jingjin.yu\}@rutgers.edu}.
This work is supported in part by NSF award IIS-1845888 and an Amazon Research Award. 
}%
}

\begin{document}

\maketitle
\thispagestyle{empty}
\pagestyle{empty}

\renewcommand{\baselinestretch}{1}
\begin{abstract}
For enabling efficient, large-scale coordination of unmanned aerial vehicles (UAVs) under the labeled setting, 
in this work, we develop the first polynomial time algorithm for the reconfiguration 
of many moving bodies in three-dimensional spaces, with provable $1.x$ asymptotic makespan optimality guarantee under high robot density. 
More precisely, on an $m_1\times m_2 \times m_3$ grid, $m_1\ge m_2\ge m_3$, our method computes solutions 
for routing up to $\frac{m_1m_2m_3}{3}$ uniquely labeled robots with uniformly randomly distributed start 
and goal configurations within a makespan of $m_1 + 2m_2 +2m_3+o(m_1)$, with high 
probability. Because the makespan lower bound for such instances is $m_1 + m_2+m_3 - o(m_1)$, 
also with high probability, as $m_1 \to \infty$, $\frac{m_1+2m_2+2m_3}{m_1+m_2+m_3}$
optimality guarantee is achieved. $\frac{m_1+2m_2+2m_3}{m_1+m_2+m_3} \in 
(1, \frac{5}{3}]$, yielding $1.x$ optimality.
In contrast, it is well-known that multi-robot path planning is NP-hard to optimally 
solve. 
In numerical evaluations, our method readily scales to support the motion planning of 
over $100,000$ robots in 3D while simultaneously achieving $1.x$ optimality. We demonstrate the application of our 
method in coordinating many quadcopters in both simulation and hardware experiments. 
%In this work, we propose a low polynomial-time algorithm for  \mpp achieving $1.x$ asymptotic optimality guarantees on solution makespan on high-dimensional grids (i.e., the time it takes to complete a reconfiguration of the robots) for random instances under very high robot density, with high probability.  
%
%
%With a linear bottleneck cost matching heuristic, the optimality of \rthddd is further improved.
% 
% Alongside the above-mentioned key result, we also establish: (1) for completely filled grids, i.e., $m_1m_2m_3$ robots, any \mpp instance may be solved in polynomial time under a makespan of $7m_1 + 14m_2$, (2) for $\frac{m_1m_2}{3}$ robots, RTH solves arbitrary \mpp instances with makespan of $3m_1+4m_2 + o(m_1)$, (3) for $\frac{m_1m_2}{2}$ robots, a variation of RTH solves a random \mpp instance with the same 1-1.5 optimality guarantee
% (4) the same $\frac{m_1+2m_2}{m_1+m_2}$ optimality guarantee holds for regularly distributed obstacles at $\frac{1}{9}$ density together with $\frac{2m_1m_2}{9}$ randomly distributed robots; such settings directly map to real-world parcel sorting scenarios.
\end{abstract}
\renewcommand{\baselinestretch}{1}

\section{Introduction}
Multi-robot path (and motion) planning (\mpp), in its many variations, has been 
extensively studied for decades 
\cite{ErdLoz86,LavHut98b,GuoPar02,blm-rvo,StaKor11,SolHal12,TurMicKum14,
SolYu15,wagner2015subdimensional,cohen2016improved,araki2017multi,tang2018complete,
wang2020walk}. 
\mpp, with the goal to effectively coordinate the motion of many robots,
finds applications in a diverse array of areas including assembly 
\cite{HalLatWil00}, evacuation \cite{RodAma10}, formation \cite{PodSuk04,SmiEgeHow08}, 
localization \cite{FoxBurKruThr00}, microdroplet manipulation \cite{GriAke05}, object transportation \cite{RusDonJen95}, search and rescue \cite{JenWheEva97}, and human 
robot interaction \cite{knepper2012pedestrian}, to list a few. 
Recently, as robots become more affordable and reliable, \mpp starts seeing
increased use in large-scale applications, e.g., warehouse automation \cite{wurman2008coordinating}, grocery fulfillment \cite{mason2019developing}, and 
UAV swarms \cite{honig2018trajectory}. 
On the other hand, due to its hardness \cite{yu2013structure,demaine2019coordinated}, 
scalable, high-performance methods for coordinating dense, large-scale robot fleets 
are scarce, especially 3D settings.  

In this work, we propose the first polynomial time algorithm for coordinating the 
motion of a large number of uniquely labeled (i.e., distinguishable) robots in 3D, with provable
$1.x$ time-optimality guarantees. 
Since \mpp in continuous domain is highly intractable \cite{hopcroft1984complexity}, 
we adapt a graph-theoretic version of \mpp and work with a $m_1\times m_2 \times m_3$
3D grid with $m_1 \ge m_2 \ge m_3$. 
\begin{figure}[t!]
\vspace{1.5mm}
        \centering
        \includegraphics[width=1\linewidth]{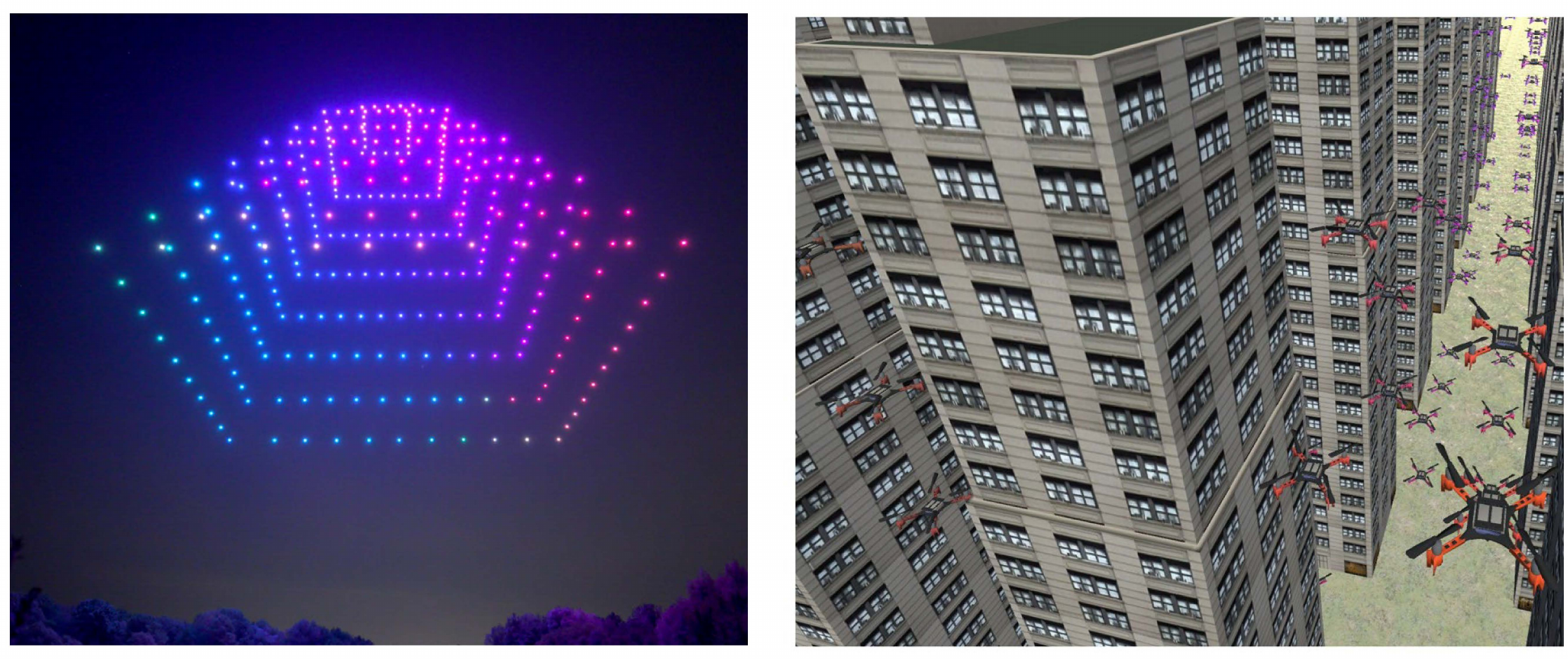}
  
        \caption{[left] A real-world drone light show where the UAVs form a 3D grid like pattern \cite{drone-flight}. [right] Our simulated large UAV swarm in the Unity environment with many tall buildings as obstacles. A video of the simulations and experiments can be found at \url{https://youtu.be/v8WMkX0qxXg}} 
        \label{fig:drones_in_cities}
        \vspace{-5mm}
\end{figure}
For up to $\frac{1}{3}$ of the grid vertices occupied with robots, which is very high, 
we perform path planning for these robots in two stages. In the first phase, we 
iteratively apply an algorithm based on the Rubik Table abstraction
\cite{szegedy2020rearrangement}  to ``shuffle'' selected dimensions of the 3D grid. 
Then, the abstract shuffle operations are translated into efficient feasible robot 
movements. 
The resulting algorithm runs in low-polynomial time and achieves high levels of 
optimality: for uniform randomly generated instances, we show that an asymptotic makespan 
optimality ratio of $\frac{m_1+2m_2+2m_3}{m_1+m_2+m_3}$ is realized, which is upper 
bounded by $\frac{5}{3}$. 
Obstacles can also be supported without significant impact on solution optimality. 
In simulation, our algorithm easily scales to graphs with over $370,000$ vertices 
and $120,000$ robots. 
%\jy{Verify the numbers} 
%
Plans computed using the two-phase process on 3D grids can be 
readily transformed into high-quality robot motion plans, e.g., 
for coordinating many aerial vehicles. 
We demonstrate our algorithm on large simulated quadcopter fleets in the Unity environment, and further demonstrate the execution of our algorithm on a fleet of $10$ Crazyflie 2.0
nano quadcopters.

\textbf{Related Work.} 
% Whereas the feasibility question has been long answered for \mpp \cite{KorMilSpi84}, due to the hardness \cite{surynek2010optimization,yu2013structure}, many methods have been developed to tackle optimally solving \mpp. 
%
The literature on multi-robot coordination is vast; we focus on graph-theoretic algorithmic and complexity results
here. 
It has been proven that solving \mpp optimally is NP-hard \cite{surynek2010optimization,yu2013structure}, even under grid-based, obstacle-free settings \cite{demaine2019coordinated}. Furthermore, it is also NP-hard to approximate within any factor less than $\frac{4}{3}$ for makespan minimization on graphs in general \cite{ma2016multi}.

Recently, many solvers have been developed to solve \mpp with decent computational efficiency and produce high-quality solutions.
% 
% Among these, combinatorial-search based solvers  have been demonstrated to be fairly effective. 
\mpp solvers can be classified as being optimal or suboptimal. 
Reduction-based optimal solvers solve the problem through reducing the \mpp problem to other problem, e.g., SAT~\cite{surynek2012towards}, answer set programming~\cite{erdem2013general}, integer linear programming (ILP)~\cite{yu2016optimal}.
Search-based optimal \mpp solvers includes EPEA* \cite{goldenberg2014enhanced}, ICTS \cite{sharon2013increasing}, CBS \cite{sharon2015conflict}, and so on.
Optimal solvers have limited scalability  due to the NP-hardness of \mpp.
Suboptimal solvers have also been extensively studied.
Unbounded solvers like push and swap~\cite{luna2011push}, push and rotate~\cite{de2014push}, windowed hierarchical cooperative A${}^*$~\cite{silver2005cooperative}, all return feasible solutions very fast at the cost 
of solution quality.
Balancing the running-time and optimality is one of the most attractive topics. 
Some algorithms emphasize the scalability without sacrificing too much optimality, e.g., ECBS~\cite{barer2014suboptimal}, DDM \cite{han2020ddm}, EECBS \cite{li2021eecbs}, PBS \cite{ma2019searching}.
Whereas these solvers achieve fairly good scalability while maintaining $1.x$-optimality in empirical evaluations, they lack provable joint efficiency-optimality guarantees. As a result, the scalability and supported robot density of these method are on the low side as compared to our method. 
There are also $O(1)-$approximate or constant factor time-optimal algorithms proposed, aiming to tackle highly dense instances, e.g.  \cite{yu2018constant,demaine2019coordinated}.
However, these algorithms only achieve low-polynomial time guarantee at the expense of huge constant factors, making them impractical.

This research builds on \cite{guo2022sub}, which addresses the \mpp problem in 2D.
In comparison to \cite{guo2022sub}, the current work describes a significant 
extension to the 3D (as well as higher dimensional) domain with many unique applications, including, e.g., the
reconfiguration of large UAV fleets, the optimal coordination of air traffic, 
and the adjustment of satellite constellations.  

%\TODO{Teng: please rewrite this section, shuffle things around a bit: you should not 
%borrow directly from even our own paper, which would be %self-plagiarism. Right now, 
%the sentences are not exactly the same, but the structure is %almost the same. You 
%don't need to make huge changes either. In general, the only thing %that can remain 
%mostly the same is the problem formulation.}

%%%%%%%%%%%%%%%%%%%%%%%%%%%%%%%%%%%%%%%%%%%%%%%%%%%%%%%%%%%%%%%%%%%%%%%%%%%%%%%%%%%
% \section{PROBLEM DEFINITION}
% \begin{problem}[Rubik Table 3D Problem \cite{szegedy2020rearrangement}]
% Let $M$ be
% an $m_1 \times m_2 \times m_3$
% table, filled with $m_1 m_2 m_3$ unique items.
% Assuming that any column/row/depth can be
% arbitrarily shuffled, given two arbitrary configurations of the
% items, $X_I$ and $X_G$, find a sequence of shuffles that takes $M$
% from $X_I$ to $X_G$.
% \end{problem}

%%%%%%%%%%%%%%%%%%%%%%%%%%%%%%%%%%%%%%%%%%%%%%%%%%%%%%%%%%%%%%%%%%%%%%%%%%%%%%%%
\section{Preliminaries}
\subsection{Multi-Robot Path Planning on 3D Grids}
We work with a graph-theoretic formulation of \mpp, which seeks collision-free
paths that efficiently route many robots on graphs. 
Specifically, consider an undirected graph $\mathcal G(V, E)$ and
$n$ robots with start configuration $S = \{s_1, \dots, s_n\}
\subseteq V$ and goal configuration 
$G = \{g_1, \dots, g_n\} \subseteq V$. A robot $i$ has start and goal vertices $s_i$ and $g_i$, respectively.
We define a {\em path} for robot $i$ as a map 
$P_i: \mathbb {N} \to V$ where $\mathbb N$ is the set of non-negative integers. 
A feasible $P_i$ must be a sequence of vertices that connect $s_i$ and $g_i$: 
1) $P_i(0) = s_i$;
2) $\exists T_i \in \mathbb N$, s.t. $\forall t \geq T_i, P_i(t) = g_i$;
3) $\forall t > 0$, $P_i(t) = P_i(t - 1)$ or $(P_i(t), P_i(t - 1)) \in E$.

%The version of \mpp we study is also known as \emph{one-shot} setting \cite{stern2019multi}.
% 
Given the 3D focus of this work, we choose the graph $\mathcal G$ to be 
a $6$-connected $m_1\times m_2 \times m_3$ grid with $m_1 \ge m_2 
\ge m_3$, as a proper discretization of the target 3D domain. Our 
aim is to minimize the \emph{makespan} of feasible routing plans, 
assuming that the transition on each edge of $G$ takes unit time. 
In other words, we are to compute a feasible path set $\{P_i\}$ that 
minimizes $\max_i\{|{P}_i|\}$.
For most settings in this work, it is assumed that the start and goal configurations are randomly generated. 
Unless stated otherwise, ``randomness'' in this paper always refers to uniform randomness. 
%
%The version of \mpp we study is sometimes referred to as the \emph{one-shot} MAPF problem \cite{stern2019multi}. 
\subsection{High-Level Reconfiguration via Rubik Tables in 3D}
Our method for solving the graph-based \mpp has two phases; the first high-level 
reconfiguration phase utilizes results on Rubik Table problems 
(\rtp)~\cite{szegedy2020rearrangement}. The Rubik Table abstraction, described 
below for the $k$-dimensional setting, $k\ge 2$, formalizes the task of carrying 
out globally coordinated token swapping operations on lattices.

\begin{problem}[{\normalfont \bf Rubik Table in $k$D (\rtpk)}]\label{p:rtkd}
Let $M$ be an $m_1 \times \ldots \times m_k$ table, $m_1\ge \ldots \ge m_k$, containing $\prod_{i=1}^km_i$ items, one in each table cell. 
In a \emph{shuffle} operation, the items in a single column in the $i$-th dimension of $M$, $1 \le i \le k$, may be permuted in an arbitrary manner. 
Given two arbitrary configurations $S$ and $G$ of the items on $M$, find a sequence of shuffles that take $M$ from $S$ to $G$.
\end{problem}

We formulate Problem~\ref{p:rtkd} differently from \cite{szegedy2020rearrangement} 
(Problem 8) to make it more natural for robot coordination tasks. For convenience, 
let the 2D and 3D version be \rtptwo and \rtp, respectively. 
A main result from \cite{szegedy2020rearrangement} (Proposition 4),
applying to the 2D setting, may be stated as follows. 
\begin{proposition}[{\normalfont \bf Rubik Table Theorem, 2D}]\label{p:rta2d}
An arbitrary Rubik Table problem on an $m_1\times m_2$ table can be solved using $m_1 + 2m_2$ column shuffles. 
\end{proposition}

Denoting the corresponding algorithm for \rtptwo as \rtatwo,
we may solve \rtp using \rtatwo as a subroutine, by treating \rtp as an \rtptwo, which is straightforward if we view a 2D slice of \rtatwo as a ``wide'' column. For example, for the $m_1 \times m_2 \times m_3$ grid, we may treat the second and third dimensions as a single dimension.
Then, each wide column, which is itself an $m_2 \times m_3$ 2D problem, can be reconfigured by applying \rtatwo.
With some proper counting, we obtain the following 3D version of 
Proposition~\ref{p:rta2d}. 

\begin{theorem}[{\normalfont \bf Rubik Table Theorem, 3D}]\label{p:rta3d}. 
An arbitrary Rubik Table problem on an $m_1\times m_2\times m_3$ table can be solved using $m_1m_2+m_3(2m_2+m_1)+m_1m_2$ shuffles. 
\end{theorem}

 \begin{figure}[h]
        \vspace{-5mm}
        \centering
        \includegraphics[width=\linewidth]{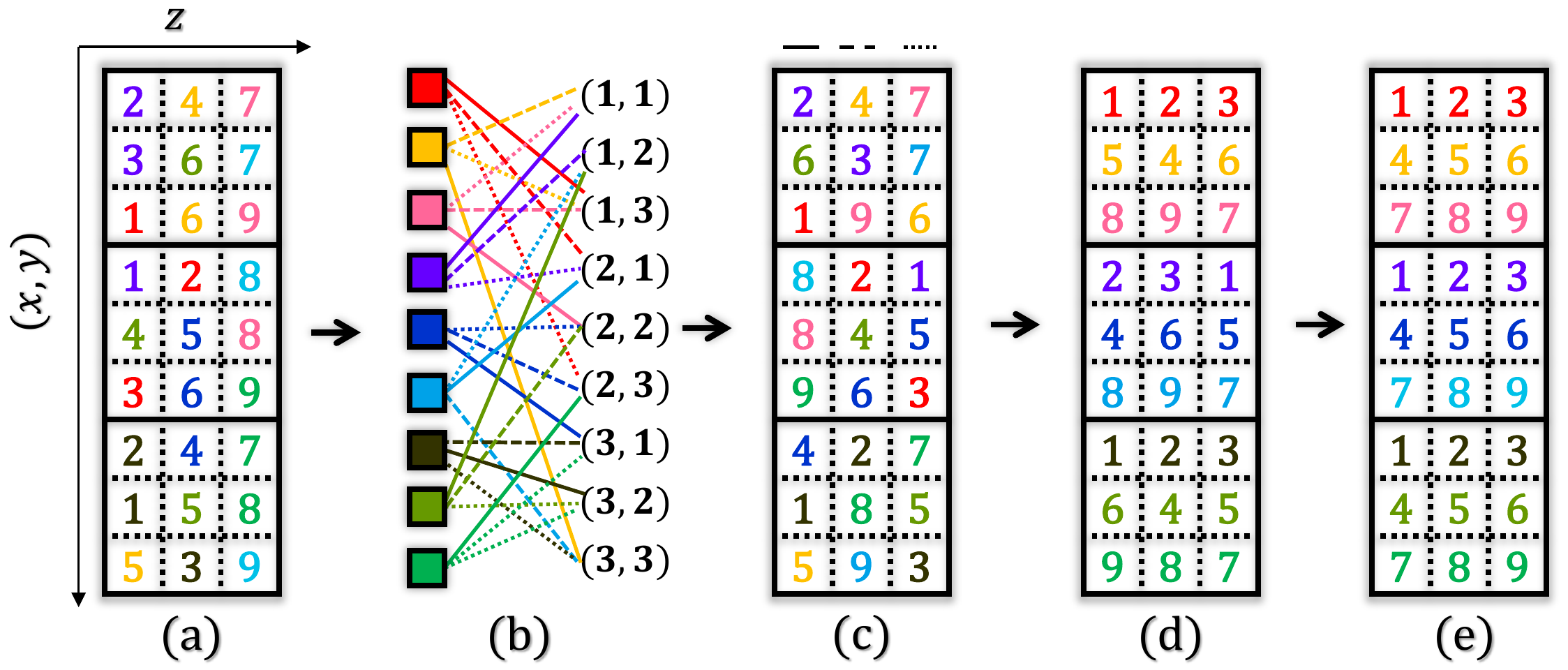}
        \vspace{-4mm}
                \caption{Illustration of applying \emph{\rta}. (a) The initial $3\times 3\times 3$ table with  a random arrangement of 27 items that are colored and labeled. A color represents the $(x,y)$ position of an item. (b) The constructed bipartite graph. The right partite set contains all the possible $(x,y)$ positions . It contains $3$ perfect matchings, determining the $3$ columns in (c). (c) Applying $z$-shuffles to (a), according to the matching results, leads to an intermediate table where each $x$-$y$ plane has one color appearing exactly once. (d) Applying wide shuffles to (c) correctly places the items according to their $(x, y)$ values (or colors). (e)  Additional $z$-shuffles fully sort the labeled items.} 
        \label{fig:rubik}
        \vspace{-2mm}
    \end{figure}

Denote the corresponding algorithm for Theorem~\ref{p:rta3d} as 
\rta, we illustrate how it works on a $3 \times 3\times 3$ table (in Fig.~\ref{fig:rubik}). \rta operates in three phases. In the first phase, a bipartite graph $B(T, R)$ is constructed based on the initial table configuration where the partite set $T$ are the colors of items representing the desired $(x,y)$ positions, and the set $R$ are the set of $(x,y)$ positions of items  (Fig.~\ref{fig:rubik}(b)). An edge is added to $B$ between $t \in T$ and $r \in R$ for every item of color $t$ in row $r$. 
From $B(T,R)$, a set of $m_1m_2$ \emph{perfect matchings} can be computed, as guaranteed by \cite{hall2009representatives}. Each matching, containing $m_3$ edges, connects all of $T$ to all of $R$, dictates how a $x$-$y$ plane should look like after the first phase. For example, the first set of matching in solid  lines in Fig.~\ref{fig:rubik}(b) says that the first $x$-$y$ plane should be ordered as the first table column shown in Fig.~\ref{fig:rubik}(c). 
After all matchings are processed, we get an intermediate table, Fig.~\ref{fig:rubik}(c). Notice that each row of Fig.~\ref{fig:rubik}(a) can be shuffled to yield the corresponding row of Fig.~\ref{fig:rubik}(c); this is the main novelty of the \rta.
After the first phase of $m_1m_2$ $z$-shuffles, the intermediate table (Fig.~\ref{fig:rubik}(c)) that can be reconfigured by applying \rtatwo with $m_1$ $x$-shuffles and $2m_2$ $y$-shuffles.
This sorts each item in the correct $x$-$y$ positions (Fig.~\ref{fig:rubik}(d)).
Another $m_1m_2$  $z$-shuffles can then sort each item to its desired $z$ position (Fig.~\ref{fig:rubik}(e)).

\section{Methods}
% 
%  In this section, we first describe a baseline adaptation of \rta to \mpp on 2D grids, as a direct combination of existing techniques. Then, we describe a highway motion primitive for significantly improving the associated makespan optimality guarantee on 2D grids, without impacting the computation time.
%  Then we demonstrate the methods applying \rta for 3D \mpp with high robot density.
%  This is followed by a matching heuristic to further improve makespan optimality.
% 
In this section, we introduce the algorithm that integrate \rta algorithm to solve \mpp on 3D grids; the built-in global coordination capability of \rtp naturally applies to solving makespan-optimal \mpp.
We assume the grid size is $m_1\times m_2\times m_3$ and there are $\frac{m_1m_2m_3}{3}$ robots.  For density less than $\frac{1}{3}$, we add virtual robots  \cite{han2018sear,yu2018constant} till reaching $\frac{1}{3}$.
\subsection{\rthddd: Adapting \rta for \mpp in 3D}
\begin{figure}[!htbp]
 \vspace{-1.5mm}
        \centering
        \includegraphics[width=0.8\linewidth]{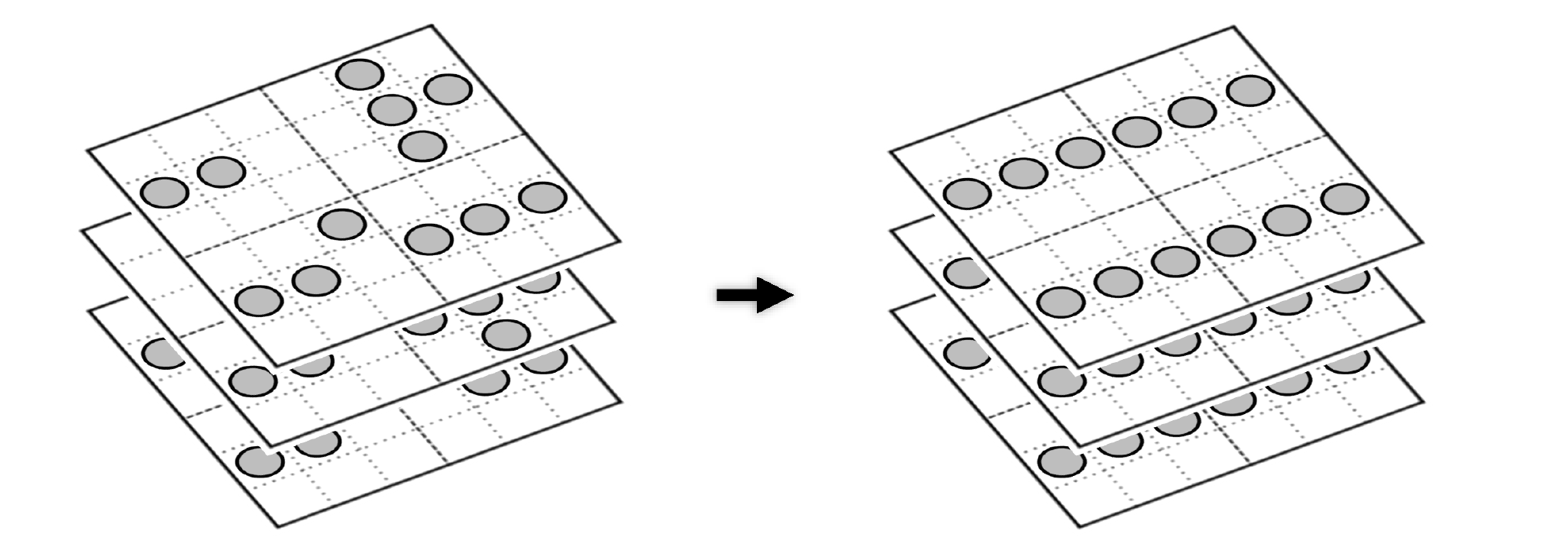}
\vspace{-2.5mm}
\caption{Applying unlabeled \mpp to convert a random configuration to a balanced centering one on  $6\times 6 \times 3$ grids.} 
        \label{fig:random_to_balanced}
\vspace{-1.5mm}
    \end{figure}

Algorithm~\ref{alg:rubik} outlines the high-level process for multi-robot routing coordination in 3D.
In each $x$-$y$ plane, we partition the grid $\mathcal G$ into $3\times 3$ cells (see, e.g.,Fig.~\ref{fig:example}). 
Without loss of generality, we assume that $m_1,m_2,m_3$ are integer multiples of 3 and first consider the case without any obstacles for simplicity.
In the first step, to make \rta applicable, we convert the arbitrary start and goal
configurations (assuming less than $1/3$ robot density)  to intermediate configurations 
where in each 2D plane, each $3 \times 3$ cell contains no more than $3$ robots (see
Fig.~\ref{fig:random_to_balanced}). 
Such configurations are called \emph{balanced configurations} \cite{guo2022sub}.
In practice, configurations are likely not ``far'' from being balanced. 
The balanced configurations $S_1,G_1$ and corresponding paths can be computed by applying any unlabeled multi-robot path planning method.
% 
% \jy{Fig. 3, the left figure has one fewer robots than the right figure, top layer.}

% Because \rta solves \rtp using column and row shuffles, they can be turned into an algorithm for \mpp if we can ``simulate'' column and row shuffles using moves allowed by \mpp problems. This is in fact achievable even when all of $\mathcal G$'s vertices are occupied by robots, by recursively applying a \emph{labeled line-swap algorithm} \cite{yu2018constant}, which can arbitrarily rearrange a line of $m$ robots embedded in a grid using $O(m)$ makespan. Such line swap operations can be carried out in parallel. 
%

% With the ability to simulate column and row swaps, we can apply \rta to \mpp on grids up to the maximum possible density. The straightforward pseudo-code is given in Alg.~\ref{alg:rubik}, with the comments in the main routine 
% \textsc{RT-MRPP} indicating the corresponding \rta phases.  
After that, \rta can be applied to coordinate the robots moving toward their intermediate goal positions $G_1$.
Function $\texttt{MatchingXY}$ finds a feasible intermediate configuration $S_2$ and route the robots to $S_2$ by simulating shuffle operations along the $z$ axis.
Function $\texttt{XY-Fitting}$ apply shuffle operations along the $x$ and $y$ axes to route each robot $i$ to its desired $x$-$y$ position  $(g_{1i}.x,g_{1i}.y)$.
In the end, function $\texttt{Z-Fitting}$ is called, routing each robot $i$ to the desired $g_{1i}$ by performing shuffle operations along the $z$ axis and concatenate the paths computed by unlabeled \mpp planner $\texttt{UnlabeledMRPP}$. 

\begin{algorithm}
\begin{small}
\DontPrintSemicolon
\SetKwProg{Fn}{Function}{:}{}
\SetKwFunction{Fampp}{UnlabeledMRPP}
\SetKwFunction{Frthddd}{RTH3D}
\SetKwFunction{Fmatchingthreed}{MatchingXY}
\SetKwFunction{Frthdd}{XY-Fitting}
\SetKwFunction{Fzfitting}{Z-Fitting}

  \caption{\rthddd \label{alg:rubik}}
%   \KwIn{robots $A=\{a, $|A| = n = m^2$}
  \KwIn{Start and goal vertices $S=\{s_i\}$ and $G=\{g_i\}$}
  \Fn{\Frthddd{$S,G$}}{
$S_1,G_1\leftarrow$\Fampp{$S,G$}\;
\Fmatchingthreed{}\;
\Frthdd{}\;
\Fzfitting{}\;
}
\vspace{1mm}
\end{small}
\end{algorithm}  
We now explain  each part of $\texttt{RTH3D}$.
$\texttt{MatchingXY}$ uses an extended version of \rta to find perfect matching that allows feasible shuffle operations. 
Here, the ``item color" of an item $i$ (robot) is the tuple $(g_{1i}.x,g_{1i}.y)$, which is the desired $x$-$y$ position it needs to go.
After finding the $m_3$ perfect matchings, the intermediate configuration $S_2$ is determined.
Then, shuffle operations along the $z$ direction can be applied to move the robots to $S_2$.
\begin{algorithm}
\begin{small}
\DontPrintSemicolon
\SetKwProg{Fn}{Function}{:}{}
\SetKwFunction{Fmatchingthreed}{MatchingXY}
  \caption{MatchingXY \label{alg:matching3d}}
%   \KwIn{robots $A=\{a, $|A| = n = m^2$}
  \KwIn{Balanced start and goal vertices $S_1=\{s_{1i}\}$ and $G_1=\{g_{1i}\}$}
  \Fn{\Fmatchingthreed{$S_1,G_1$}}{
$A\leftarrow[1,...,n]$\;
$\mathcal{T}\leftarrow$ the set of $(x,y)$ positions in $S_1$\;
\For{$(r,t)\in \mathcal{T}\times \mathcal{T}$}{
    \If{$\exists i\in A$ where $(s_{1i}.x,s_{1i}.y)=r\wedge (g_{1i}.x,g_{1i}.y)=t$}{
        add edge $(r,t)$ to $B(T,R)$\;
        remove $i$ from $A$ \;
    }

}
    compute matchings $\mathcal{M}_1,...,\mathcal{M}_{m_3}$ of $B(T, R)$\;
     $A\leftarrow [1,...,n]$\;
 \ForEach{$\mathcal{M}_c$ and $(r,t)\in \mathcal{M}_c$}{
 \If{$\exists i\in A$ where $(s_{1i}.x,s_{1i}.y)=r\wedge (g_{1i}.x,g_{1i}.y)=t$}{
 $s_{2i}\leftarrow (s_{1i}.x, s_{1i}.y,c)$ and remove $i$ from $A$\;
  mark robot $i$ to go to $s_{2i}$\;
 }
 }
   perform simulated $z$-shuffles in parallel \;
  
}
\vspace{1mm}
\end{small}
\end{algorithm}
The robots in each $x$-$y$ plane will be reconfigured by applying $x$-shuffles and $y$-shuffles.
We need to apply \rtatwo for these robots in each plane, as demonstrated in Algorithm \ref{alg:rth2d}.
In $\texttt{RTH2D}$, for each 2D plane,  the ``item color" for robot $i$ is its desired $x$ position $g_{1i}.x$.
For each plane, we compute the $m_2$ perfect matchings to determine the intermediate position $g_{2}$.
Then each robot $i$ moves to its $g_{2i}$ by applying $y$-shuffle operations.
In Line 18, each robot is routed to its desired $x$ position by performing $x$-shuffle operations.
In Line 19, each robot is routed to its desired $y$ position by performing $y$-shuffle operations.
\begin{algorithm}
\begin{small}
\DontPrintSemicolon
\SetKwProg{Fn}{Function}{:}{}
\SetKwFunction{Fxyfitting}{XY-Fitting}
\SetKwFunction{Frthdd}{RTH2D}
  \caption{XY-Fitting \label{alg:rth2d}}
%   \KwIn{robots $A=\{a, $|A| = n = m^2$}
  \KwIn{Current positions $S_2$ and balanced goal positions $G_1$}
  \Fn{\Fxyfitting{}}{
  \For{$z\leftarrow[1,...,m_3]$}{
$A\leftarrow \{i|s_{2i}.y=z\}$\;
\Frthdd{$A,z$}\;
}
}
  \Fn{\Frthdd{$A,z$}}{

$\mathcal{T} \leftarrow$ the set of $x$ positions of $S_2$\;
\For{$(r,t)\in \mathcal{T}\times\mathcal{T}$}{
       \If{$\exists i\in A$ where $s_{2i}.x=r\wedge g_{1i}.x=t$}{
       \If{robot $i$ is not assigned}{
       add edge $(r,t)$ to $B(T,R)$\;
        mark $i$ assigned \;
       }

    }
      compute matchings $\mathcal{M}_1,...,\mathcal{M}_{m_2}$ of $B(T, R)$\;
}
$A'\leftarrow A$\;
    %   $A\leftarrow \{i|s_{2i}.y=z\}$\;
   \ForEach{$\mathcal{M}_c$ and $(r,t)\in \mathcal{M}_c$}{
 \If{$\exists i\in A'$ where $s_{2i}.x=c\wedge g_{1i}.x=t$}{
 $g_{2i}\leftarrow (s_{2i}.x, c,z)$ and remove $i$ from $A'$\;
  mark robot $i$ to go to $g_{2i}$\;
 }
 }
  route each robot $i\in A$ to $g_{2i}$\;
  route each robot $i\in A$ to  $(g_{2i}.x,g_{1i}.y,z)$\;
  route each robot $i\in A$ to  $(g_{1i}.x,g_{1i}.y,z)$\;
}

\vspace{1mm}
\end{small}
\end{algorithm} 

After all the robots reach the desired $x$-$y$ positions, another round of $z$-shuffle operations in $\texttt{Z-Fitting}$ can route the robots to the balanced goal configuration computed by an unlabeled \mpp planner.
In the end, we concatenate all the paths as the result.

\subsection{Efficient Shuffle Operation with High-Way Heuristics}
\begin{figure}[htbp]
        \centering
        \includegraphics[width=\linewidth]{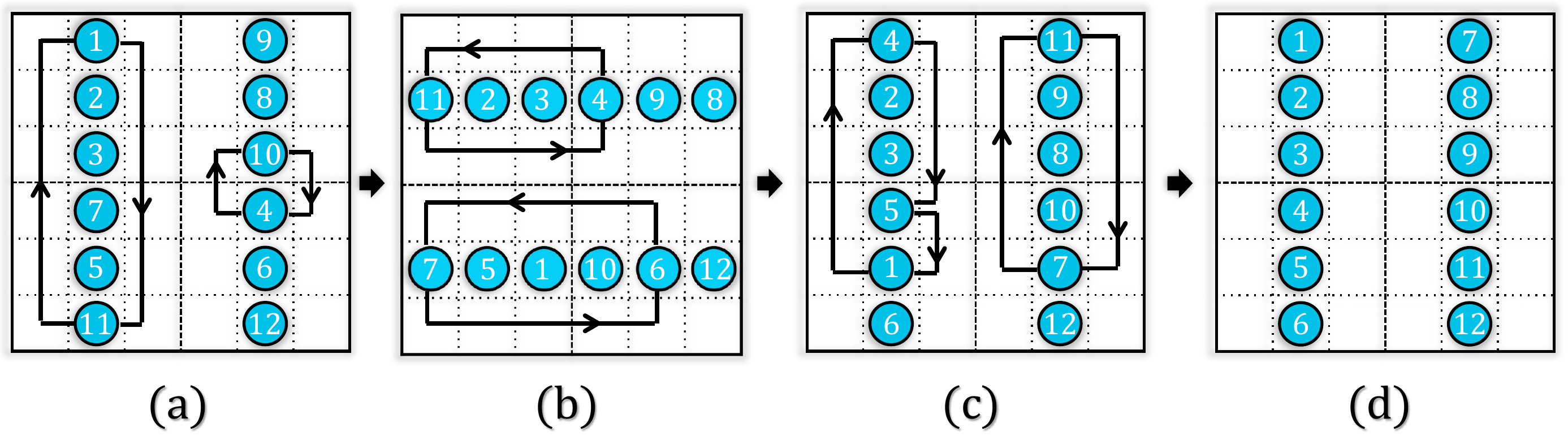}
\vspace{-2mm}        
        \caption{An example of applying \rthdd to solve an instance. (a) The initial configuration; (b) The intermediate configuration obtained from matching;   (d) The desired configuration;} 
        \label{fig:example}
\vspace{-2mm}     
% \tg{the figure has no problem if downloading the pdf}
    \end{figure}

In this section, we explain how to simulate the shuffle operation exactly.
We use the shuffle operation in $x$-$y$ plane as an example.
% 
% Whereas \rtmapf runs in polynomial time and provides constant factor makespan optimality in expectation, 
% the constant factor is very large due to the extreme density. 
% In practice, an robot density of $1/3$ (of the vertices of $\mathcal G$) is already very high. For instances with robot density no more than $1/3$, as it turns out, the constant factor can be dropped significantly by employing a ``highway'' motion primitive to simulate the shuffle operation.  
% %
We partition the grid $\mathcal G$ into $3\times 3$ cells (see, e.g., Fig.~\ref{fig:example}). 
Without loss of generality, we assume that $m$ is an integer multiple of 3 and first consider the case without any obstacles for simplicity.
% \tg{Maybe we can add some sentences to explain this is only for simplicity concerns. Wider columns are trivial.  we can even use two columns for the shuffle operations. }

% We use Fig.~\ref{fig:example}, where Fig.~\ref{fig:example}(a) is a random start configuration and Fig.~\ref{fig:example}(f) is a random goal configuration, as an example to illustrate the new algorithm which we call Rubik Table with Highways, or \rthdd.
We use Fig.~\ref{fig:example}, where Fig.~\ref{fig:example}(a) is the initial start configuration and Fig.~\ref{fig:example}(d) is the desired goal configuration in $\texttt{MatchingXY}$, as an example to illustrate how to simulate the shuffle operations.
%
% \rthdd involves two phases: \emph{unlabeled reconfiguration} and \emph{\mpp resolution with Rubik Table and highway heuristics}. In the first phase, arbitrary start and goal configurations (assuming less than $1/3$ robot density) are converted to intermediate configurations where each $3 \times 3$ cell contains no more than $3$ robots. We call such configurations \emph{balanced configurations}. In practice, configurations are likely not ``far'' from being balanced. 
%
% In the example, this corresponds to Fig.~\ref{fig:example}(a)$\to$Fig.~\ref{fig:example}(b) and Fig.~\ref{fig:example}(f)$\to$Fig.~\ref{fig:example}(e) (note that \mpp solutions are time-reversible). We call configurations like  Fig.~\ref{fig:example}(b)-(e), which have all robots concentrated vertically or horizontally in the middle of the $3\times 3$ cells, \emph{centered balanced configurations} or simply \emph{centered configurations}. 
% %
% Completing the first phase requires solving two unlabeled \mpp problems \cite{Ma2016OptimalTA}, easily doable in polynomial time. 
%
% \tg{the figure looks fine if downloading the pdf}
In $\texttt{MatchingXY}$, the initial configuration is a vertical ``centered'' configuration (Fig.~\ref{fig:example}
(a)).
\rta is applied with a highway heuristic to get us from Fig.~\ref{fig:example}(b) to Fig.~\ref{fig:example}(c), transforming between vertical centered configurations and horizontal centered configurations. 
To do so, \rta is applied (e.g., to Fig.~\ref{fig:example}) to obtain two intermediate configurations (e.g., Fig.~\ref{fig:example}(b)(c)).
To go between these configurations, e.g., Fig.~\ref{fig:example}(b)$\to$Fig.~\ref{fig:example}(c), we apply a heuristic by moving robots that need to be moved out of a $3\times 3$ cell to the two sides of the middle columns of Fig.~\ref{fig:example}(b), depending on their target direction. If we do this consistently, after moving robots out of the middle columns, we can move all robots to their desired goal $3\times 3$ cell without stopping nor collision. 
%This is what we mean by the high-way heuristics; the re-configuration instance is well-formed \cite{Ma2017LifelongMP} because each path found by using high-ways will not traverse starts and goals of other robots. 
Once all robots are in the correct $3\times 3$ cells, we can convert the balanced configuration to a centered configuration in at most $3$ steps, which is necessary for carrying out the next simulated row/column shuffle. 
Adding things up, we can simulate a shuffle operation using no more than $m + 5$ steps. 
 
% Putting everything together, \rthdd takes no more than $7m + o(m)$ steps to complete and can be much lower, translating to an optimality ratio of $3.5x$ in expectation. We establish this more formally in the next section along with other properties of the algorithm. 

% \subsection{\rthddd}
% We now demonstrate how to apply \rta on 3D grids for 1/3 robot density.
% % 
% We first  consider the $x$-$y$ plane as a ``wide column"  of size $m_1m_2$.
% % 
% Then the problem is reduced to a $m_1m_2\times m_3$ 2D \mpp problem.
% % 
% Directly applying the \rthdd yields two depth-shuffles and one ``wide column" shuffle. 
% % 
% For simulating the ``wide column" shuffle, one can further apply \rthdd if each plane has $m_1m_2/3$ robots.
% % 
% We can ensure this in the unlabeled \mpp phase.
% % 
% That is we can first apply unlabeled \mpp to turn the start/goal configuration to the configuration where each plane has $m_1m_2/3$ robots and these robots are at the balanced configuration.
% % 
% This has another benefit that we do not need unlabeled \mpp reconfigurations when applying \rthdd to simulate the ``wide column" shuffles.
\subsection{Improving Solution Quality via Optimized Matching}
The matching steps in \rta determine all the intermediate positions and therefore have great impact on solution optimality.
% The optimality of Rubik Table  method is determined by the first preparation phase. 
% The matchings determine the intermediate states in all three phases. 
%
Finding arbitrary perfect matchings is fast but can be further optimized. 
We replace the matching method from finding arbitrary perfect matchings to finding a min-cost matching in both of $\texttt{MatchingXY}$ and $\texttt{RTH2D}$. 

%\textcolor{red}{For this part, to keep subscripts consistent with what we have before, let's use $i$ instead of $a$ to index robots, $c$ instead of $i$ to index columns, and $t$ intead of $j$ to index robot's color/type. Also update column to $c$ in Eq. (3)}

The matching heuristic we developed is based on \emph{linear bottleneck assignment (LBA)}, which runs in polynomial time. A 2D version of the heuristic is introduced in \cite{guo2022sub}, to which readers are referred to for more details. 
Here, we illustrate how to apply LBA-matching in $\texttt{MatchingXY}$.
For the matching assigned to height $z$, the edge weight of the bipartite graph is computed greedily.
If a vertical column $c=(x,y)$ contains robots of ``color" $t$, here $t=(g_2.x,g_2.y)$, we add an edge $(c,t)$ and its edge cost is
\begin{equation}
    % C_{ij}=\min_{a.g_y=i}C_{ra}(\lambda=0)
    C_{ct}=\min_{(g_{2i}.x,g_{2i}.y)=t}C_{zi}(\lambda=0).
\end{equation}
We choose $\lambda=0$ to optimize the first phase. Optimizing the third phase ($\lambda=1$) would give similar results.
After constructing the weighted bipartite graph, an $O(m^{2.5}/\log m)$ LBA algorithm \cite{burkard2012assignment} is applied to get a minimum bottleneck cost matching for height $z$. Then we remove the assigned robots and compute the next minimum bottleneck cost matching for next height. 
After getting all the matchings $\mathcal{M}_z$, we can further use LBA to assign $\mathcal{M}_z$ to a different height $z'$ to get a smaller makespan lower bound. The cost for assigning matching $\mathcal{M}_z$ to height $z'$ is defined as 
\begin{equation}
    C_{\mathcal{M}_zz'}=\max_{i\in\mathcal{M}_z}C_{z'i}(\lambda=0).
\end{equation}
% The total time-complexity of using LBA heuristic for matching is $O(m^{3.5}/\log m)$.
The same approach can also be applied to $\texttt{RTH2D}$. We denote \rthddd with LBA heuristics as  \rthdddlba.

%%%%%%%%%%%%%%%%%%%%%%%%%%%%%%%%%%%%%%%%%%%%%%%%%%%%%%%%%%%%%%%%%%%%%%%%%%%%%%
\section{Theoretical Analysis}
First, we introduce two important lemmas about \emph{well-connected} grids which have been proven in \cite{yu2018constant}.
\begin{lemma}\label{lemma: anonymous_mkpn}
On  an $m_1\times m_2 \times m_3$ grid,  any unlabeled \mpp can be solved using $O(m_1+m_2+m_3)$ makespan. 
\end{lemma}
\begin{lemma}\label{lemma: makespan}
On an $m_1\times m_2\times m_3$ grid, if the underestimated makespan of an \mpp instance is $d_g$ (usually computed by ignoring the inter-robot collisions), then this instance can be solved using $O(d_g)$ makespan. 
\end{lemma}
% 
% \begin{lemma}[Rubik Rectangle]
% An $m_1\times m_2$  rubik rectangle  problem can be solved in $2m_1+n_2$ shuffles with deterministic time complexity being $O(m_1^2m_2)$ or expected time complexity  being $O(m_2m_1\log m_1)$.
% \end{lemma}
% 
We proceed to analyze the time complexity of \rthddd on $m_1\times m_2\times m_3$ grids.
% \begin{proposition}[Time Complexity of \rthddd]
% \rthddd runs in $O(nV^2)$.
% \end{proposition}
% \begin{proof}
The dominant parts of \rthddd are computing the perfect matchings and solving unlabeled \mpp.
First we analyze the running time for computing the  matchings.
Finding $m_3$ ``wide column" matchings runs in $O(m_3 m_1^2m_2^2)$ deterministic time or $O(m_3m_1m_2\log(m_1m_2))$ expected time.
We apply \rthdd for simulating ``wide column" shuffle, which requires $O(m_3 m_1^2m_2)$  deterministic time or $O(m_1m_2m_3\log m_1)$ expected time. 
Therefore, the total time complexity for Rubik Table part is $O(m_3 m_1^2m_2^2+m_1^2m_2m_3)$.
For unlabeled \mpp, we can use the max-flow based algorithm \cite{yu2013multi} to compute the makespan-optimal solutions. 
The max-flow portion can be solved in $O(n|E|T)=O(n^2(m_1+m_2+m_3))$  where $|E|$ is the number of edges and $T=O(m_1+m_2+m_3)$ is the time horizon of the time expanded graph \cite{ford1956maximal}.
Note that any unlabeled \mpp can be applied, for example,  the algorithm in \cite{yu2012distance} is distance-optimal but with convergence time guarantee. %\cite{yu2013multi}.
% 

% \end{proof}
Note that the choice of 2D plane can also be $x$-$z$ plane or $y$-$z$ plane.
% .
In addition, one can also perform two ``wide column" shuffles plus one $z$ shuffle, which yields $2(2m_1+m_2)+m_3$ number of shuffles and $O(m_1m_2m_3^2+m_3m_1m_2^2)$. 
This requires more shuffles but shorter running time.

Next, we derive the optimality guarantee.

\begin{proposition}[Makespan Upper Bound]
\rthddd returns  solution with worst makespan $3m_1+4m_2+4m_3+o(m_1)$.
\end{proposition}
\ifarxiv
\begin{proof}
The Rubik Table portion has a makespan of $(2m_3+2m_2+m_1)+o(m_1)$.
For the unlabeled \mpp portion, we use $m_1+m_2+m_3$, which is the maximum grid distance between any two nodes on the $m_1 \times m_2 \times m_3$ grid, as a conservative makespan upper bound.
Therefore, the total makespan upper bound is  $2(m_1+m_2+m_3)+(2m_3+2m_2+m_1)+o(m_1)=3m_1+4m_2+4m_3+o(m_1).$
When $m_1=m_2=m_3$ for cubic grids, it  turns out to be $11m_3+o(m_3).$
\end{proof}
\fi
We note that the unlabeled \mpp upper bound is actually a very  conservative estimation. 
In practice, for such dense instances, the unlabeled \mpp usually requires much fewer steps.
Next, we analyze the makespan when start and goal configurations are uniformly randomly generated based on the well-known \emph{minimax grid matching} result.
% \begin{theorem}[Minimax Grid Matching \cite{leighton1989tight}]\label{t:minimax}
% Consider a $\sqrt{N}\times \sqrt{N}$ square grid containing $N$ points following the uniform distribution. 
% Let $\ell$ be the minimum length such that there exists a perfect matching of the $N$ points to the grid points in the square for which the distance between every pair of matched points is at most $\ell$. Then $\ell = O(\log^{3/4}N)$.
% \end{theorem}
\begin{theorem}[Multi-dimensional Minimax Grid Matching \cite{shor1991minimax} ]\label{t:minimax3d}
For $k\geq 3,$
consider $N$ points following the uniform
distribution in $[0,N^{1/k}]^k$. 
Let $\mathcal{L}$ be the minimum length such that there exists
a perfect matching of the $N$ points to the grid points  that are regularly spaced in the
$[0,N^{1/k}]^k$ for which the distance between every pair of matched
points is at most $\mathcal{L}$. Then $\mathcal{L} = O(\log^{1/k}N)$ with high probability.
\end{theorem}

\begin{proposition}[Asymptotic Makespan]\label{p:expected_makespan}
\rthddd returns solutions with $m_1+2m_2+2m_3+o(m_1)$ asymptotic makespan for \mpp instances with $\frac{m_1m_2m_3}{3}$ random start and goal configurations on 3D grids, with high probability.
Moreover, if $m_1=m_2=m_3$, \rthddd returns $5m_3+o(m_3)$ makespan-optimal solutions.
\end{proposition}
\ifarxiv
\begin{proof}
% To establish the result, all we need to show is that given $\eta\xi m^3/3$ robots on an $\xi m \times \eta m\times m$ grid, we can take a random initial configuration to a new configuration such that there are three robots in each $3 \times 3\times 3$ cell.
% % 
% This can be proved using a relatively direct application of Theorem~\ref{t:minimax3d}; 
% we proceed by working with $N = m^2/9$ for three times (i.e., we partition the $\xi \eta m^3/3$ points into three batches, each of which is also uniformly random). 
% % 
% Applying Theorem~\ref{t:minimax} to an $\frac{m}{3}\times \frac{m}{3}$ grid and $m^2/9$ points says that the Euclidean distance for a minimum length perfect matching that moves the points to grid points is $\ell = O(\log^{1/3}m)$.
% % 
% Considering the actual grid having a side length of $m$ instead of $m/3$ and the distance in our setting is Manhattan only introduce some small constant factors. 
% % 
% Therefore, to move $m^2/3$ robots so that each $3\times 3$ cell contains three robots, the minimum matching distance is $O(\log^{3/4}m) = o(m)$.
% % 
% 
First, we point out that the theorem of minimax grid matching (Theorem~\ref{t:minimax3d}) can be generalized to any rectangular cuboid as the matching distance mainly depends on the discrepancy of the start and goal distributions, which does not depend on the shape of the grid. 
Using the minimax grid matching result, the matching distance from a random configuration to a centering configuration, which is also the underestimated makespan of unlabeled \mpp, scales as $O(\log^{1/3}m_1)$.
Using the Lemma \ref{lemma: makespan}, it is not difficult to see that the matching obtained this way can be readily turned into an unlabeled \mpp plan without increasing the maximum per robot travel distance by much, which remains at $o(m)$.
Therefore, the asymptotic makespan of \rthddd is $m_1+2m_2+2m_3+O(\log^{1/3}m_1)=m_1+2m_2+2m_3+o(m_1)$ with high probability.
If $m_1=m_2=m_3$, the asymptotic makespan  is $5m_3+o(m_3)$, with high probability.
\end{proof}
\fi
\begin{proposition}[Asymptotic Makespan Lower Bound]\label{p:asymptotic_lowerbound}
For \mpp instances on $ m_1\times  m_2\times m_3$ grids with $\Theta(m_1m_2m_3)$ random start and goal configurations on 3D grids, the makespan lower bound is asymptotically approaching $m_1+m_2+m_3$, with high probability.
\end{proposition}
\ifarxiv
\begin{proof}
We examine two opposite corners of the $m_1\times  m_2\times m_3$ grid. 
At each corner, we examine an $\alpha  m_1\times \alpha  m_2\times \alpha m_3$ sub-grid for some constant $\alpha \ll 1$. 
The Manhattan distance between a point in the first sub-grid and another point in the second sub-grid is larger than $(1-6\alpha)(m_1+m_2+m_3)$. 
As $m\to \infty$, with $\Theta(m^3)$ robots,  a start-goal pair falling into these two sub-grids
 can be treated as an event following binomial distribution $B(k,\alpha^6)$ when $m\rightarrow \infty$.
 The probability of having at least one success trial among $n$ trials is $p=1-(1-\alpha^6)^n$, which goes to one when $m\rightarrow \infty$.

Because $(1 - x)^y < e^{-xy}$ for $0 < x < 1$ and $y > 0$,
%\footnote{This is because $\log(1-x) < -x$ for $0 < x < 1$; multiplying both sides by a positive $y$ and exponentiate with base $e$ then yield the inequality.}, 
$p > 1 - e^{-\alpha^6n}$. Therefore, for arbitrarily small $\alpha$, we may choose $m_1$ such that $p$ is arbitrarily close to $1$. 
The probability of a start-goal pair falling into these two sub-grids is asymptotically approaching one 
, meaning that the makespan goes to $(1-6\alpha)(m_1+m_2+m_3)$. 
\end{proof}
\fi

\begin{corollary}[Asymptotic Makespan Optimality Ratio]
\rthddd yields asymptotic $1+\frac{m_2+m_3}{m_1+m_2+m_3}$ makespan optimality ratio for \mpp instances with $\Theta(m_1m_2m_3)\le \frac{m_1m_2m_3}{3}$ random start and goal configurations on 3D grids, with high probability.
\end{corollary} 
\ifarxiv
\begin{proof}
If $n < \frac{m_1m_2m_3}{3}$, we add virtual robots with randomly generated start and use the same for the goal, until we reach $n =\frac{m_1m_2m_3}{3}$ robots, which then allows us to invoke Proposition~\ref{p:expected_makespan}.
In viewing Proposition~\ref{p:expected_makespan} and Proposition~\ref{p:asymptotic_lowerbound}, solution computed by \rthddd guarantees a makespan optimality ratio of $\frac{m_1+2m_2+2m_3}{m_1+m_2+m_3}=1+\frac{m_2+m_3}{m_1+m_2+m_3}$ as $m_3\to \infty$.
Moreover, if $m_1=m_2=m_3$, \rthddd returns $\frac{5}{3}$ makespan-optimal solution.
\end{proof}
\fi
% \emph{Flat Rubik Table} problem is a variant of Rubik Table 3D problem with the size of one of the dimension is fixed as a constant (i.e. $m_3=K$.)
% \begin{proposition}
% For an $\xi m\times \eta m\times K$ Flat Rubik Table problem with 1/3 robot density, the \rthddd algorithm yields $2\xi m+(2K+\eta m)=(2\xi+\eta)m+2K$ makespan solution.  
% \end{proposition}

\begin{corollary}[Asymptotic Optimality, Fixed Height]\label{c:fh}
For an \mpp on an $ m_1\times  m_2\times K$ grid, $m_1>m_2\gg K$, and $\frac{1}{3}$ robot density, the \rthddd algorithm yields $1+\frac{m_1}{m_1+m_2}$ optimality ratio, with high probability.
If $m_1=m_2$, the asymptotic makespan optimality ratio is $1.5$.
\end{corollary}

\begin{theorem}
 Consider an $k$-dimensional cubic grid with grid size $m$. If robot density is less than $1/3$ and start and goal configurations are uniformly distributed, generalizations to \rta can solve the instance with asymptotic makespan optimality being $\frac{2^{k-1}+1}{k}$.
\end{theorem}
\ifarxiv
\begin{proof}
By theorem \ref{t:minimax3d}, the unlabeled \mpp takes $o(m)$ makespan (note for $k=1,2$, the minimax grid matching distance is still $o(m)$~\cite{leighton1989tight}).
Extending proposition \ref{p:asymptotic_lowerbound} to $k$-dimensional grid, the asymptotic lower bound is $mk$.
We now prove that the asymptotic makespan $f(k)$ is $(2^{k-1}+1)m+o(m)$ by induction.
The Rubik Table algorithm solves a $d$-dimensional problem by using two 1-dimensional shuffles and one $(k-1)$-dimensional ``wide column" shuffle.
Therefore, we have $f(k)=2m+f(k-1)$.
It's trivial to see $f(1)=m+o(m),f(2)=3m+o(m)$, which yields that $f(k)=2^{k-1}m+m+o(m)$ and makespan optimality ratio being $\frac{2^{k-1}+1}{k}$.
\end{proof}
\fi
%%%%%%%%%%%%%%%%%%%%%%%%%%%%%%%%%%%%%%%%%%%%%%%%%%%%%%%%%%%%%%%%%%%%%%%%%%%%%%%%
\section{Simulations And Experiments}
In this section, we evaluate the performance of our polynomial time, asymptotic 
optimal algorithms and compare them with fast and near-optimal solvers, 
\ecbs($w$=1.5) \cite{barer2014suboptimal} and ILP with 16-split heuristic \cite{yu2016optimal,guo2021spatial}. 
For the unlabeled \mpp planner, we use the algorithm \cite{yu2012distance}. Though it minimizes total distance, we find it is much faster than max-flow method \cite{yu2013multi} and the makespan optimality is very close to the optimal one on well-connected grids.
All the algorithms are implemented in C++.
We mention that, though not presented here, we also evaluated push-and-swap
\cite{luna2011push}, which yields good computation time but substantially worse 
optimality (with ratio $> 25$) on the large and dense settings we attempt. 
We also examined prioritized methods, e.g., \cite{ma2019searching,silver2005cooperative},
which faced significant difficulties in resolving deadlocks. 
All experiments are performed on an Intel\textsuperscript{\textregistered} Core\textsuperscript{TM} i7-9700 CPU at 3.0GHz. Each data point is an average over 
20 runs on randomly generated instances, unless otherwise stated.
A running time limit of $300$ seconds is imposed over all instances. 
The optimality ratio is estimated as compared to conservatively estimated  
lower bound.

In addition to numerical evaluations, we further demonstrate using \rta-LBA to 
coordinate many UAVs in the Unity \cite{juliani2018unity} environment  as well as to plan trajectories for $10$ Crazyflie 2.0 nano quadcopter in 3D. 
We will make our source code available upon the publication of this 
work. 

\subsection{Evaluations on 3D Grids}
\begin{figure}[htbp]
%\vspace{-1.5mm}
        \centering
        \includegraphics[width=1\linewidth]{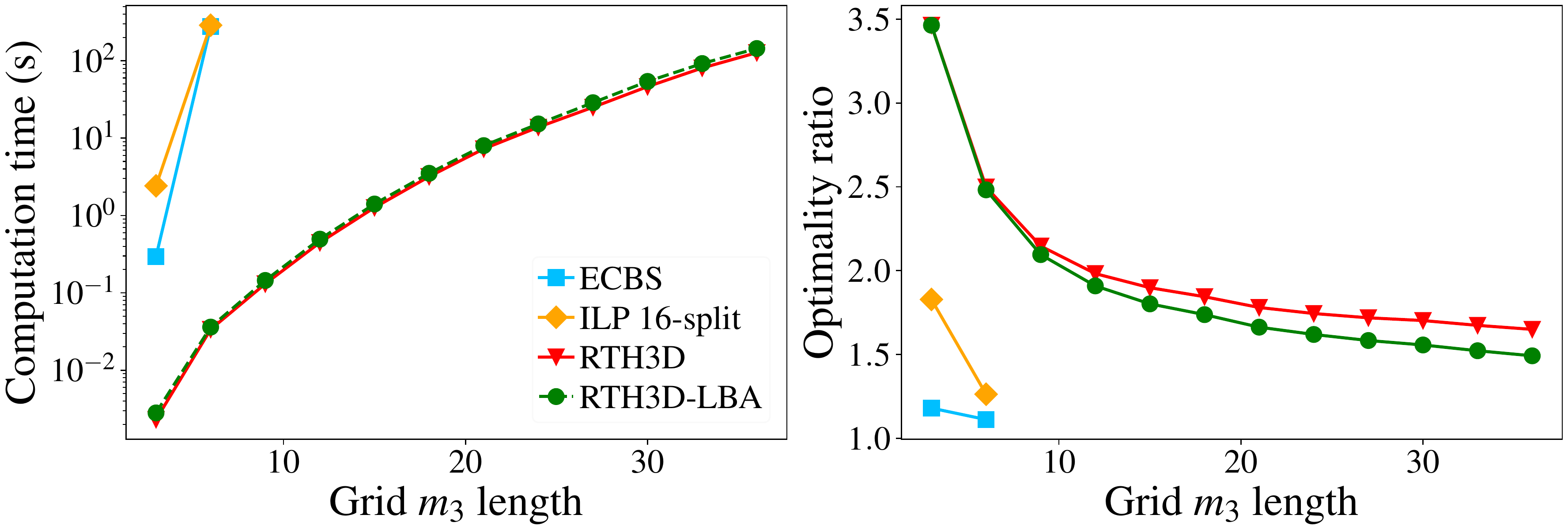}
% \vspace{-1.5mm}
\caption{Computation time and optimality ratio on grids with varying grid size and $m_1:m_2:m_3=4:2:1$.} 
        \label{fig:random_rec}
\vspace{-1.5mm}
    \end{figure}

In the first evaluation, we fix the aspect ratio $m_1:m_2:m_3=4:2:1$ and $\frac{1}{3}$ robot density,  and examine the performance \rta methods on obstacle-free grids with varying size.
Start and goal configurations are randomly generated; the results are shown in
Fig.~\ref{fig:random_rec}.
\ilp with 16-split heuristic and \ecbs computes solution with better optimality ratio but have poor scalability.
In contrast, \rthddd and \rthdddlba readily scale to grids with over $370,000$ vertices and $120,000$ robots.
Both of the optimality ratio of \rthddd and \rthdddlba decreases as the grid size increases, asymptotically approaching $\sim 1.7$ for \rthddd and $\sim 1.5$ for \rthdddlba.

\begin{figure}[htbp]
\vspace{-1.5mm}
        \centering
        \includegraphics[width=1\linewidth]{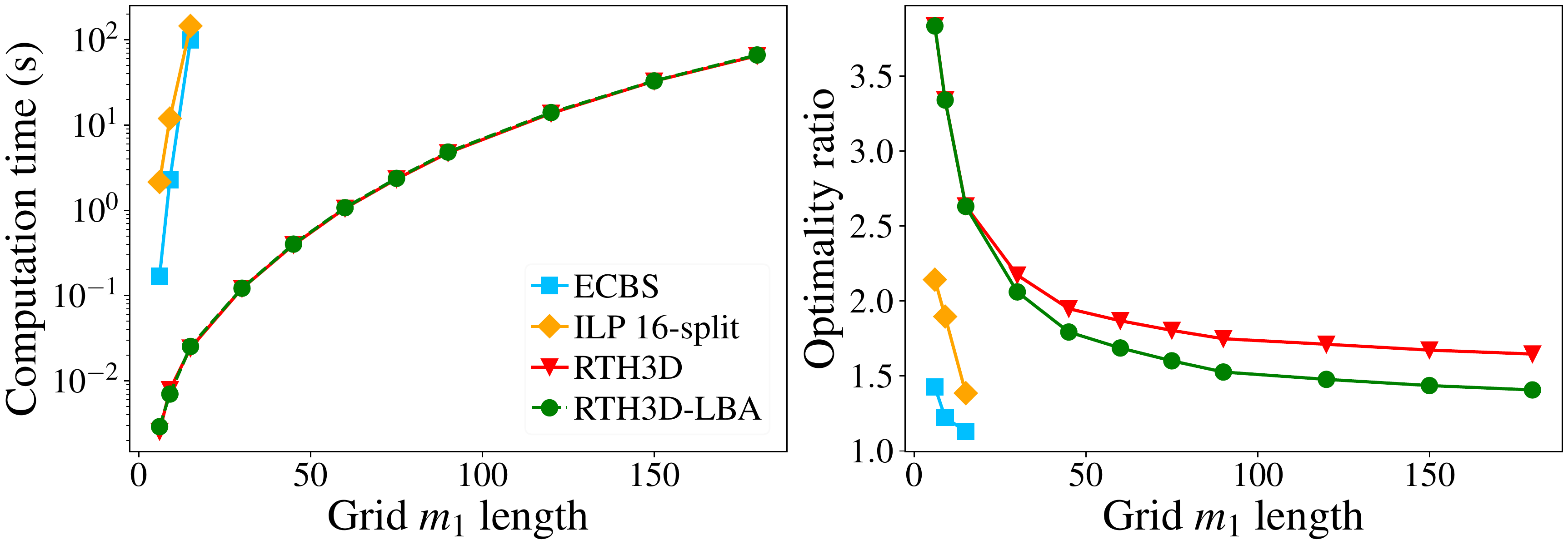}
% \vspace{-1.5mm}
\caption{Computation time and optimality ratio on grids with $m_1:m_2=1$ and $m_3=6$.} 
        \label{fig:flat}
\vspace{-1.5mm}
    \end{figure}

In a second evaluation, we fix $m_1:m_2=1:1, m_3=6$ and robot density at $\frac{1}{3}$, 
and vary  $m_1$.
As shown in Fig.~\ref{fig:flat}, \rthddd and \rthdddlba show exceptional scalability and the asymptotic 1.5 makespan optimality ratio, as predicted by Corollary~\ref{c:fh}.
    
\begin{figure}[htbp]
\vspace{-1.5mm}
        \centering
        \includegraphics[width=1\linewidth]{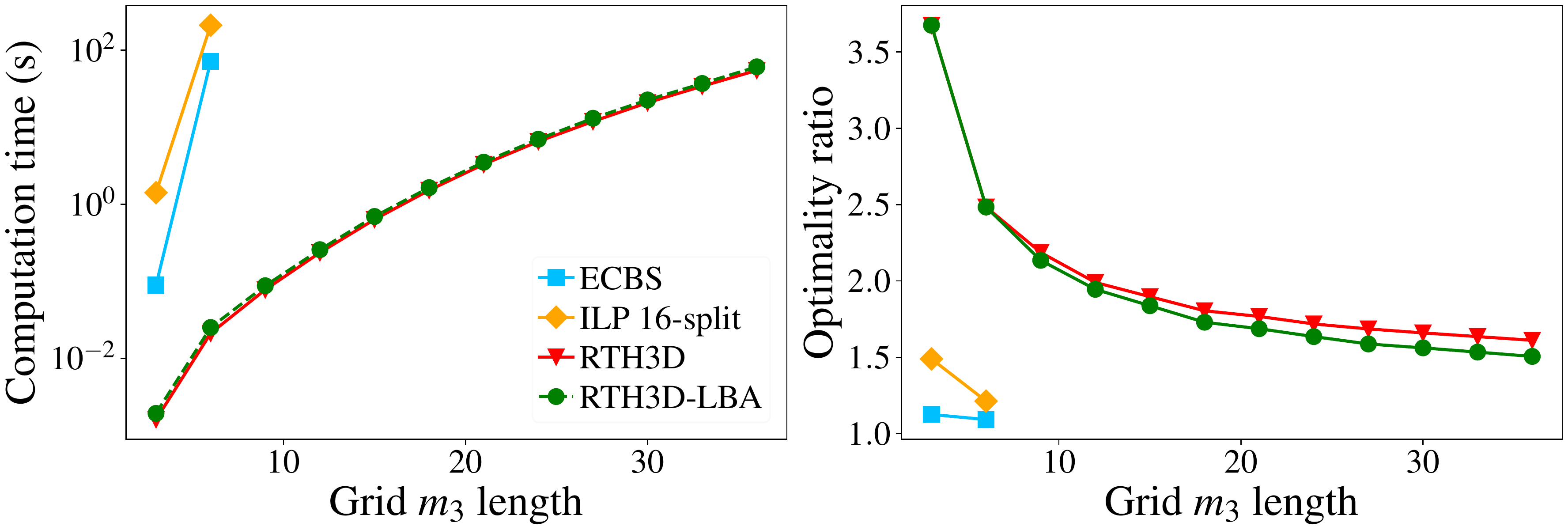}
% \vspace{-1.5mm}
\caption{Computation time and optimality ratio on 3D environments with obstacles. $m_1:m_2:m_3=4:2:1$.} 
        \label{fig:obs_rec}
\vspace{-1.5mm}
    \end{figure}

    \rthddd and \rthdddlba  support scattered obstacles where obstacles are small and regularly distributed.
As a demonstration of this capability, we simulate setting with an environment containing many tall buildings, a snapshot of which is shown  Fig. \ref{fig:drones_in_cities}.
These ``tall building'' obstacles are located at positions $(3i+1,3j+1)$, which 
corresponds to an obstacle density of over $10\%$.
UAV swarms have to avoid colliding with the tall buildings and are not allowed to fly higher than those buildings.
We fix the aspect ratio at $m_1:m_2:m_3=4:2:1$ and robot density at $\frac{2}{9}$.
The evaluation result is shown in Fig.~\ref{fig:obs_rec}.

\subsection{Impact of Robot Density}
\begin{figure}[htbp]
\vspace{-1.5mm}
        \centering
        \includegraphics[width=1\linewidth]{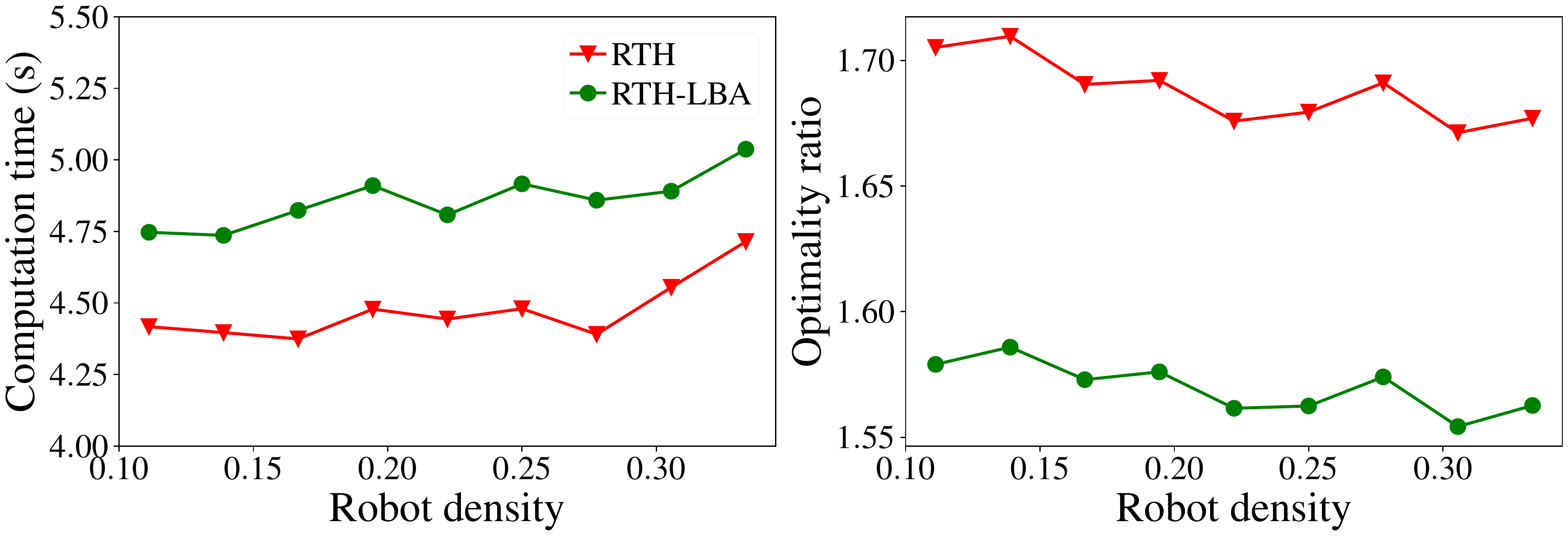}
\caption{Computation time and optimality ratio on a $120\times 60\times 6$ grid with varying robot density.} 
        \label{fig:robot_density}
\vspace{-1.5mm}
    \end{figure}

Next, we vary robot density, fixing the environment as a $120\times 60\times 6$ grid.
For robot density lower than $\frac{1}{3}$, we add virtual robots with random start and goal configurations for perfect matching computation. 
When applying the shuffle operations and evaluating optimality ratios, virtual robots are removed.
Therefore, the optimality ratio is not overestimated.
The result is plotted in Fig.~\ref{fig:robot_density}, showing that robot density has little 
impact on the running time. 
On the other hand, as robot density increases, the lower bound and the makespan required by unlabeled \mpp are closer to theoretical limits. Thus, the makespan optimality ratio is actually better.

\subsection{Special Patterns}
\begin{figure}[htbp]
\vspace{-1.5mm}
        \centering
        \includegraphics[width=1\linewidth]{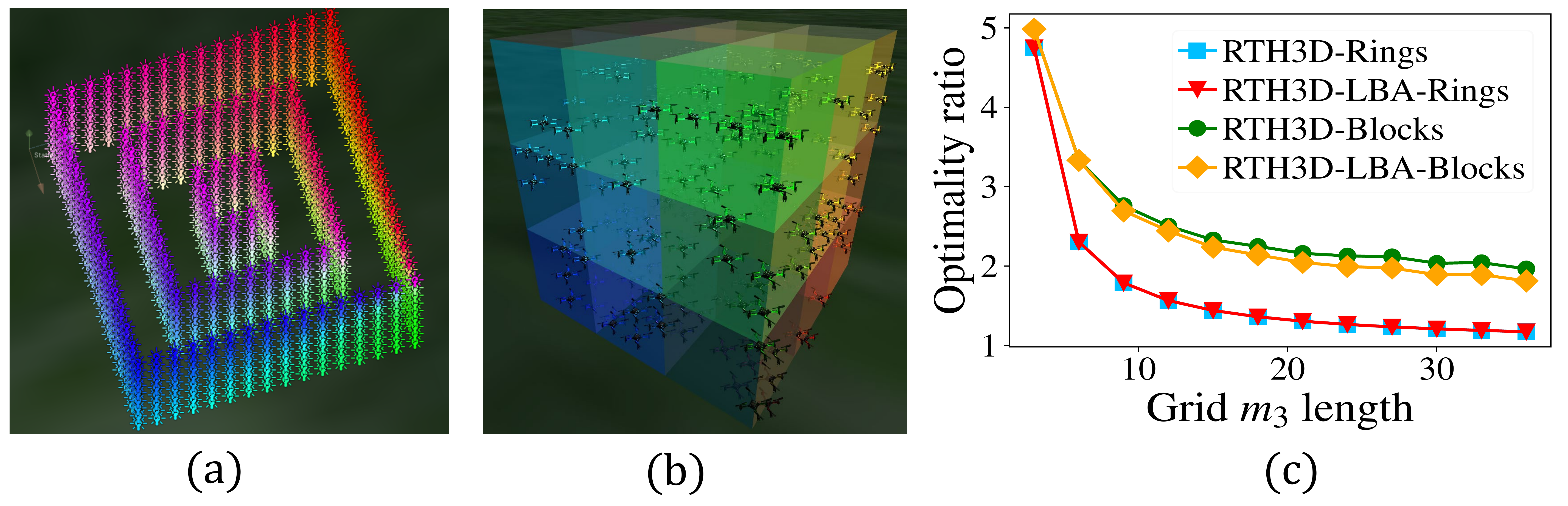}
\caption{Special patterns and the associated optimality ratios.} 
        \label{fig:special_pattern}
\vspace{-1.5mm}
    \end{figure}

In addition to random instances, we also tested instances with start and goal configurations forming special patterns, e.g., 3D ``ring'' and ``block'' structures (Fig.~\ref{fig:special_pattern}).
For both settings, $m_1 = m_2=m_3$.
In the first setting, ``rings'', robots form concentric square rings in each $x$-$y$ plane. Each robot and its goal are centrosymmetric.
In the ``block" setting, the grid is divided to $27$ smaller cubic blocks. The robots in one block need to move to another random chosen block.
Fig.~\ref{fig:special_pattern}(c) shows the optimality ratio results of both settings on grids with varying size. Notice that the optimality ratio for the ring approaches $1$ for large environments.

\subsection{Crazyswarm Experiment}
\begin{wrapfigure}[6]{r}{1.2in}
\vspace{-6.5mm}
  \begin{center}
    \includegraphics[width=1.2in]{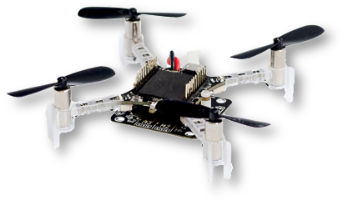}
  \end{center}
  \vspace{-4mm}
  \caption{Crazyflie 2.0 nano quadcopter.}
  \label{fig:cf2}
\end{wrapfigure}
\begin{figure}[!htbp]
\vspace{1mm}
        \centering
        \includegraphics[width=1\linewidth]{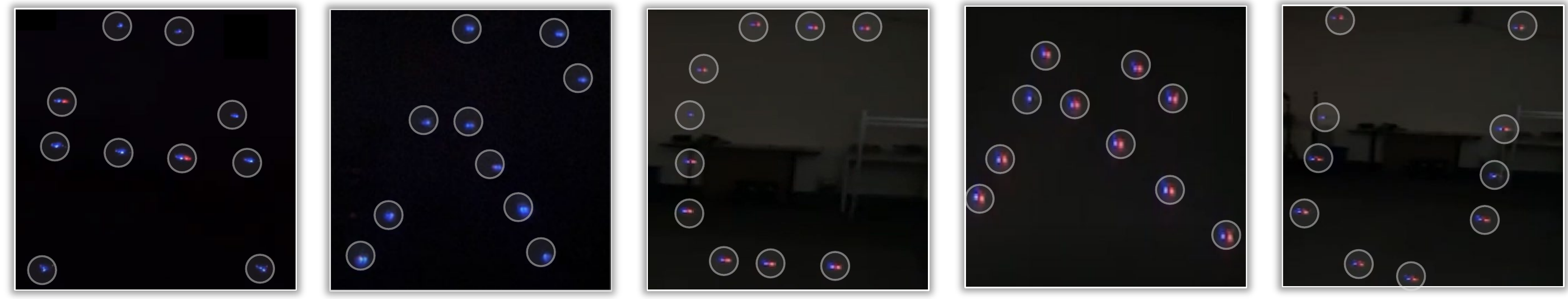}
\caption{A Crazyswarm \cite{preiss2017crazyswarm} with 10 Crazyflies following paths computed by \rta on $6\times6\times 6$ grids, transitioning between letters ARC-RU. The figure shows five snapshots during the process. We note that some letters are skewed in the pictures which is due to non-optimal camera angles when the videos were taken.} 
        \label{fig:real_drone}
\vspace{-1.5mm}
    \end{figure}

Paths planned by \rta can also be readily transferred to real UAVs.
To 
demonstrate this, $10$ Crazyflie 2.0 nano quadcopters are choreographed to form the ``A-R-C-R-U'' pattern, letter by letter, on $6\times 6 \times 
6$ grids (Fig.~\ref{fig:real_drone}).
The discrete paths are computed by \rthddd.
Due to relatively low robot density,  shuffle operations are further 
optimized to be more efficient.
Continuous trajectories are then computed  based on \rthddd plans by 
applying the method described in \cite{honig2018trajectory}.
%%%%%%%%%%%%%%%%%%%%%%%%%%%%%%%%%%%%%%%%%%%%%%%%%%%%%%%%%%%%%%%%%%%%%%%%%%%%%%
\section{Conclusion}
In this study, we proposed to apply Rubik Table results \cite{szegedy2020rearrangement} to solve \mpp. 
Combining \rta, efficient shuffling, and matching heuristics, we obtain 
a novel polynomial time algorithm, \rthddd-LBA, that is provably $1.x$
makespan-optimal with up to $\frac{1}{3}$ robot density. 
In practice, our methods can solve problems on graphs with over $300,000$ 
vertices and $100,000$ robots to $1.5$ makespan-optimal. 
%
%To our knowledge, no previous \mpp solvers provide dual guarantees on 
%low-polynomial running time and practical optimality. 
%
Paths planned by \rthddd-LBA readily translate to high quality 
trajectories for coordinating large number of UAVs and are demonstrated 
on a real UAV fleet. 

%In this work, we presented the  Rubik Table based algorithm with high-way motion primitive to solve large and dense \mpp problems on grids or sorting-center like maps where obstacles are regularly spaced.
%We demonstrate that the algorithm runs in polynomial time and has $O(1)$ makespan optimality with reasonably small constant factor.

In future work, we plan to expand in several directions, enhancing both the 
theoretical guarantees and the methods' practical applicability. 
In particular, we are interested in planning better paths that carefully 
balance between the nominal solution path optimality at the grid level and 
the actual path optimality during execution, which requires the consideration 
of the dynamics of the aerial vehicles that are involved. We will also 
address domain-specific issues for UAV coordination, e.g., down-wash, possibly
employing learning-based methods \cite{shi2020neural}.
\bibliographystyle{IEEEtran}
% %\bibliography{references}
{\footnotesize\bibliography{all}}
%%%%%%%%%%%%%%%%%%%%%%%%%%%%%%%%%%%%%%%%%%%%%%%%%%%%%%%%%%%%%%%%%%%%%%%%%%%%%%%%

\end{document}